\pdfoutput=1

\documentclass[11pt]{article}

\usepackage[]{acl}

\usepackage{times}
\usepackage{latexsym}

\usepackage[T1]{fontenc}

\usepackage[utf8]{inputenc}

\usepackage{microtype}

\usepackage{amsmath}
\usepackage{algorithm,algpseudocode}
\usepackage{amsfonts}
\usepackage{graphicx}
\usepackage{booktabs}
\usepackage{amssymb}
\usepackage{pifont}
\usepackage{tikz}
\usepackage{enumitem}
\usepackage{xcolor}

\newcommand{\Fref}[1]{Figure~\ref{#1}}

\newcommand{\Tref}[1]{Table~\ref{#1}}

\title{CMU's IWSLT 2024 Simultaneous Speech Translation System}

\author{Xi Xu\thanks{~ Equal contribution.} ~ Siqi Ouyang\footnotemark[1] ~ Brian Yan ~ Patrick Fernandes ~ \\ \textbf{William Chen ~  Lei Li ~ Graham Neubig ~ Shinji Watanabe }\\
Language Technologies Institute, Carnegie Mellon University, USA ~~\\ 
\texttt{\{xixu, siqiouya\}@cs.cmu.edu}
}

\begin{document}
\maketitle
\begin{abstract}
This paper describes CMU's submission to the IWSLT 2024 Simultaneous Speech Translation (SST) task for translating English speech to German text in a streaming manner. Our end-to-end speech-to-text (ST) system integrates the WavLM speech encoder, a modality adapter, and the Llama2-7B-Base model as the decoder. We employ a two-stage training approach: initially, we align the representations of speech and text, followed by full fine-tuning. Both stages are trained on MuST-c v2 data with cross-entropy loss. We adapt our offline ST model for SST using a simple fixed hold-n policy. Experiments show that our model obtains an offline BLEU score of 31.1 and a BLEU score of 29.5 under 2 seconds latency on the MuST-C-v2 tst-COMMON.
\end{abstract}

\section{Introduction}

This paper presents CMU's submission to the IWSLT 2024 \cite{iwslt:2024} Simultaneous Speech Translation (SST) task, focusing on streaming English speech to German text translation. Recent advancements in large language models (LLMs) have demonstrated their potential to be a strong backbone for offline ST~\cite{huang2023speech,zhang2023tuning}. In this year's submission, we build an end-to-end offline ST model with WavLM \cite{Chen_2022} and Llama2-7B-Base \cite{touvron2023llama} following the practice of LST \cite{zhang2023tuning}. Then we adapt the offline model for simultaneous translation. 

We prepare our end-to-end ST model in the following steps:
\begin{enumerate}
    \item Offline ST with WavLM and Llama2-7B-base.
    \item Online adaptation of offline model via hold-n policy and incremental beam search.
\end{enumerate}

\section{Task Description}

The IWSLT 2024 SST track\footnote{\url{https://iwslt.org/2024/simultaneous}} English-German direction is a shared task for streaming speech-to-text translation of English TED talks. The task requires the system to generate the translation without modifying its previous outputs. The average lagging (AL) \cite{ma-etal-2019-stacl} of SST systems must be below 2 seconds on MuST-C v2.0 tst-COMMON set \cite{di-gangi-etal-2019-must}. Note that AL has been modified from its original definition \cite{ma-etal-2020-simuleval}.

Following the constraint of data and pretrained weights, we use MuST-C v2.0 as the only training set and leverage pretrained models of WavLM and Llama2-7B-Base.

\begin{figure*}[!ht]
    \centering
    \includegraphics[width=0.75\linewidth]{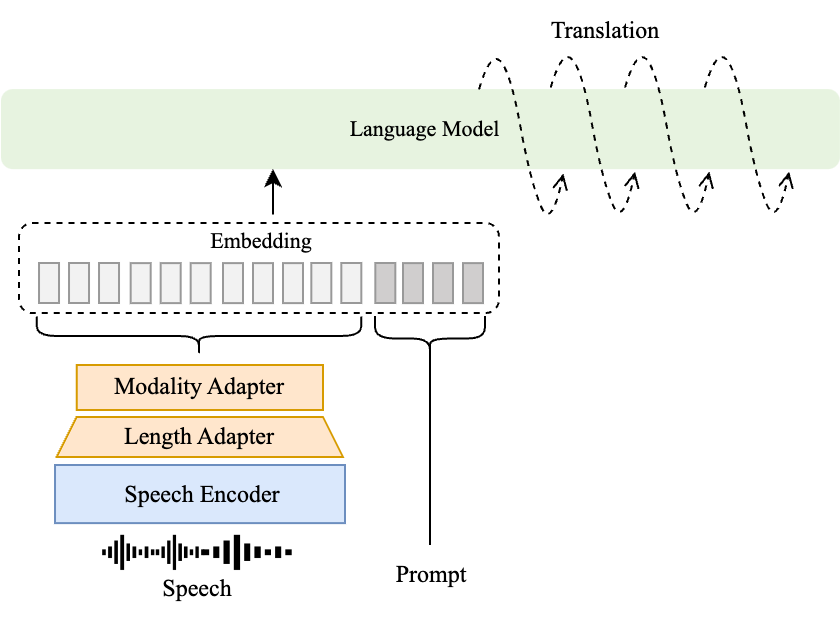}
    \caption{Offline ST model architecture based on WavLM encoder and Llama2 7B decoder.}
    \label{fig:offline_model}
\end{figure*}

\section{System Description}



As shown in \Fref{fig:offline_model}, our offline ST models consists of three primary components: a speech encoder, an adapter, and a LLM decoder. 

For the speech encoder, we employ the WavLM model \footnote{\url{https://huggingface.co/microsoft/wavlm-large}}, which has been pre-trained on 94,000 hours data including LibriLight \cite{librilight}, VoxPopuli \cite{wang-etal-2021-voxpopuli} and GigaSpeech \cite{gigaspeech}. We use the output of last encoder layer as the speech representation.

The modality adapter consists of two components: a length adapter and a modality adapter. The length adapter consists of two 1-dimensional convolutional layers with a kernel size of 5, a stride size of 2, padding of 2, and a hidden size of 1024. The modality adapter is a linear layer that projects the output of the length adapter to the embedding space of LLM. 

We use Llama2-7B-Base as the LLM decoder. The LLM decoder takes the output of the modality adapter and autoregressively generate the target text translation.

\subsection{Offline Speech Translation (ST)}
\label{sec:st}

For each sample, given speech \( X^S \), the reference translation \( X^T \), and the prompt \( X^P \), we initially transform the speech signal into a feature representation via the speech encoder:
\begin{equation}
H^S = \text{Encoder}(X^S),
\end{equation}
where \( H^S = [h^S_1, \ldots, h^S_T] \) with $T$ denoting the sequence length of the feature representation. To reconcile the length difference between the speech feature sequence \(H^S\) and its corresponding text, we downsample the speech with a length adapter. 

To clarify further, the length adapter transforms \(H^S\) using a pair of 1-dimensional convolutional layers, which can be represented as:
\begin{equation}
Z^S = \text{Length adapter}(H^S; k, s, p, h),
\end{equation}
where k is the kernel size, s is the stride, p is the padding, and h denotes the number of convolutional filters. 
The reduced temporal dimension is  \( Z^S = [z^S_1, \ldots, z^S_N] \), where
\begin{equation}
N = \left\lfloor \frac{T - k + 2p}{s} \right\rfloor + 1,
\end{equation}
Next, a projector is applied to transform the speech features \( Z^S \) into \( E^S \) with the same dimension as the LLM input embedding. We use a single hidden layer as the projector,
\begin{equation}
E^S = \text{Linear}(Z^S).
\end{equation}
Finally, we feed the speech embedding \( E^S \), translation embedding \( E^T \), and prompt embedding \( E^P \) into the template to compose the final input \( E \) of LLM,
\begin{equation}
E^T = \text{Emb}(\text{Tokenizer}(X^T)),
\end{equation}
\begin{equation}
E^P = \text{Emb}(\text{Tokenizer}(X^P)),
\end{equation}
\begin{equation}
E =
\begin{cases}
\text{Template}(E^S, E^P, E^T) & \text{if training}, \\
\text{Template}(E^S, E^P, \Tilde{E}^T) & \text{if inference},
\end{cases}
\end{equation}
where $\text{Emb}$ is the LLM embedding layer, $\Tilde{E}^T$ is the embedding of model's previously generated tokens. 

The template is formatted as:
\begin{quote}
    \textit{$<$P$>$} \textit{USER}: \textit{$<$S$>$} \textit{ASSISTANT}:\textit{$<$T$>$}
\end{quote}
where \textit{$<$P$>$} represents the system prompt\footnote{We use the following system prompt: "You are a large language and speech assistant. You are able to understand the speech content that the user provides, and assist the user with a variety of tasks using natural language. Follow the instructions carefully and explain your answers in detail."}, \textit{$<$S$>$} denotes the speech embedding, and \textit{$<$T$>$} is the target reference or generated translation.

\begin{table*}[t]
  \centering
    \resizebox {.8\linewidth} {!} {
\begin{tabular}{lcccc}
\toprule

\textsc{Model} & \textsc{Quality} & \multicolumn{3}{c}{\textsc{Latency}} \\
\midrule
\textsc{Offline Speech Translation (ST)} & \textsc{SacreBLEU $\uparrow$} &  \textsc{AL $\downarrow$} & \textsc{LAAL $\downarrow$} &\textsc{LAAL\_CA} $\downarrow$\\
\cmidrule(lr){1-1}\cmidrule(lr){2-5}
WavLM-LLaMA2 (Ours) & 31.1 & 5.85 & 5.85 & 7.09\\

\midrule
\textsc{Simul Speech Translation (SST)} & \textsc{SacreBLEU $\uparrow$} & \textsc{AL $\downarrow$} & \textsc{LAAL $\downarrow$} &\textsc{LAAL\_CA} $\downarrow$\\
\cmidrule(lr){1-1}\cmidrule(lr){2-5}
WavLM-LLaMA2-AlignAtt \cite{Papi_2023} & 27.8 & 2.00 & 2.21 & 2.93\\
WavLM-LLaMA2 (Ours) & 29.5 & 1.96 & 2.22 & 3.16\\


\bottomrule

\end{tabular}
}

    \caption{Results of our English to German ST/SST models on MuST-C-v2 tst-COMMON. Latency for offline ST is calculated using a wait-k policy with k set to infinity.}
    \label{tab:main}
\end{table*}

We finetune our offline ST model following a 2-stage strategy. In the first stage, we finetune the speech encoder together with the adapters, while keeping the LLM frozen. In the second stage, we finetune the entire model. We employ cross entropy loss in both stages. In addition, we apply rule-based filtering \cite{ouyang-etal-2022-impact} of the dataset to clean the unnecessary speaker names from the training set.

\subsection{Simultaneous Speech Translation (SST)}
\label{sec:sst}

\begin{algorithm}[!t]
\caption{Selective Output of Speech Chunk Hypotheses}
\label{alg:selective_output}
\footnotesize
\begin{algorithmic}[1]
\Procedure{SelectiveOutput}{$\operatorname{hyps}, n$}
\State $\operatorname{prunedHyps} = \{\}$
\For {$c \in \{1, \ldots, C\}$}
    \State $W^{(c)} = \operatorname{hyps}[c]$ 
    \State $l = |W^{(c)}|$
    \If {$\operatorname{source\_finished}$}
        \State $\operatorname{prunedHyps}[c] = W^{(c)}$
    \Else
        \State $n' = \min(n, l)$ 
        \State $W^{(c)}_{\text{prefix}} = W^{(c)}_{0:l-n'}$
        \If {$W^{(c)}_{\text{prefix}}$ is not empty}
            \State $\operatorname{prunedHyps}[c] = W^{(c)}_{\text{prefix}}$
        \Else
            \State $\operatorname{action} = \text{Read}$
            \State \textbf{break}
        \EndIf
    \EndIf
\EndFor
\State \Return $\operatorname{prunedHyps}$
\EndProcedure
\end{algorithmic}
\end{algorithm}

We adapt our offline ST model for streaming inference using hold-n policy.
Our scheme uses a fixed duration (e.g. 2 seconds) to compute the encoder representations on chunks of input speech.
With each new chunk, we re-compute the encoder representations using the entire given input speech.

As shown in Algorithm 1, for each chunk $c$, we obtain the corresponding hypotheses $W^{(c)}$ using beam search given partial speech input. We then determine the number of tokens $n'$ to withhold based on the minimum of the predefined value $n$ and the length of the current chunk's hypotheses $l$. The prefix $W^{(c)}_{\text{prefix}}$ is obtained by selecting the tokens from index 0 to $l-n'$.

\section{Experimental Setup}


We use the AdamW optimizer with a cosine learning rate decay and a warmup ratio of 0.2.  The learning rate commences at 2e-4 for the first training stage and is reduced to 2e-5 for the second stage. We train the first stage for 6 epochs and train the second stage for 1 epoch.

We employ an early stopping strategy with a patience of 6 epochs, evaluating every 1000 steps in Stage 1 and every 200 steps in Stage 2. The batch size is set to 128 for both stages. All models are trained on 4 Nividia A6000 GPUs with Deepspeed's ZeRO training strategy. The training times for the first and second stages are approximately 29 hours and 9 hours, respectively. We select the checkpoints with the lowest dev loss for testing.

For offline testing, we use a beam size of 4 to generate translations. In the simultaneous testing scenario, we set the start seconds to 2, indicating the initial wait time before processing speech chunks. We employ a hold-n strategy with n set to 7, meaning that the last 7 tokens of each chunk are withheld until more context is available. The beam size is set to 4, and the chunk size is set to 2500ms.



We evaluate translation quality using SacreBLEU \cite{post-2018-call}. We evaluate translation latency for SST with average lagging (AL) \cite{simuleval2020} and length-adaptive average lagging (LAAL) \cite{papi2022over} using SimulEval toolkit \cite{simuleval2020}.

\section{Results}
\label{sec:res}

\Tref{tab:main} shows the quality and latency of our SST system as measured on En-De tst-COMMON.
We also include the offline ST performance of our model for reference. We implement the Alignatt policy \cite{Papi_2023} as a baseline for our model, we set start seconds to 2, speech segment size to 1000ms. We set number of frames to 20 and use attention from all layers of the LLM decoder with greedy decoding.


\begin{table}[t]
    \centering
    \setlength\tabcolsep{4pt}
    \begin{tabular}{lcccc}
        \toprule
        & \multicolumn{2}{c}{\textbf{Wav2vec}} & \multicolumn{2}{c}{\textbf{WavLM}} \\
        \cmidrule(lr){2-3} \cmidrule(lr){4-5}
        \textbf{LLM} & Stage1 & Stage2 & Stage1 & Stage2 \\
        \midrule
        TowerInstruct & - & - & 29.64 & - \\
        Tower & 29.35 & 30.53 & 30.11 & 31.64 \\
        LLaMA2 & - & 30.02 & 25.50 & 30.31 \\
        \bottomrule
    \end{tabular}
    \caption{SacreBLEU score of different Speech Encoder and LLMs, all models are trained on the original MuST-C 2.0 data without data cleaning.}
    \label{tab:ablation}
\end{table}

From ST to SST, we observe a 5\% quality degradation (31.1 to 29.5 SacreBLEU). However, this comes with significant latency improvements. The Average Lagging (AL) decreases from 5.85 to 1.96 seconds, a 66.5\% reduction. The Length Adaptive Average Lagging (LAAL) improves from 5.85 to 2.22 seconds, a 62.1\% decrease.

We also investigate the impacts of different LLMs and speech encoders, as shown in \Tref{tab:ablation}. We compare WavLM with a CTC fine-tuned Wav2vec 2.0 large model\footnote{\url{https://dl.fbaipublicfiles.com/fairseq/wav2vec/wav2vec\_vox\_960h\_pl.pt}}. This Wav2vec model was pre-trained on 53.2k hours of untranscribed speech from LibriVox and fine-tuned on 960 hours of transcribed speech from Librispeech, as well as on pseudo-labels. Our results show that replacing Wav2vec with WavLM yields a significant improvement: a 1.1 BLEU score increase when using the Tower LLM \cite{alves2024tower} as the decoder, and a 0.3 BLEU score increase with LLaMA2 as the decoder. This suggests that the performance gains from a well-pretrained speech encoder are more pronounced when coupled with LLMs of higher translation capability.

Our analysis of the performance between different LLMs used as decoders shows that the Tower LLM\footnote{\url{https://huggingface.co/Unbabel/TowerBase-7B-v0.1}}, subjected to continued pre-training on a curated multilingual dataset of 20 billion high-quality tokens, exhibits a marked performance advantage over LLaMA2 in the initial stage of training. However, during the second stage, when the LLM backend is trainable, Tower quickly overfits, implying potential overlap between the MuST-C corpus and the data involved in Tower's pretraining. Tower Instruct\footnote{\url{https://huggingface.co/Unbabel/TowerInstruct-7B-v0.1}}, which undergoes supervised fine-tuning (SFT) on instruction dataset for various translation-related tasks, achieves a slightly lower BLEU score compared to the base model. To mitigate overfitting during the second stage of training with Tower, a reduced learning rate of 7e-6 is used, compared to the 2e-5 learning rate applied to LLaMA2 training.

\section{Conclusion}

In this paper, we describe the submission of CMU's English to German simultaneous speech-to-text translation systems for the IWSLT 2024 Simultaneous track.
We start by building a offline speech-to-text system which leverages self-supervised speech and text foundation models. 
We then adapt this offline model for streaming inference, enabling simultaneous speech-to-text translation.

\section*{Acknowledgements}
Siqi Ouyang is supported by an Amazon Research Award. 
This work used the Extreme Science and Engineering Discovery Environment (XSEDE) ~\cite{xsede}, which is supported by National Science Foundation grant number ACI-1548562; specifically, the Bridges system ~\cite{nystrom2015bridges}, as part of project cis230075p, which is supported by NSF award number ACI-1445606, at the Pittsburgh Supercomputing Center.

\bibliography{anthology,custom}

\begin{thebibliography}{19}
\expandafter\ifx\csname natexlab\endcsname\relax\def\natexlab#1{#1}\fi

\bibitem[{Alves et~al.(2024)Alves, Pombal, Guerreiro, Martins, Alves, Farajian, Peters, Rei, Fernandes, Agrawal, Colombo, de~Souza, and Martins}]{alves2024tower}
Duarte~M. Alves, José Pombal, Nuno~M. Guerreiro, Pedro~H. Martins, João Alves, Amin Farajian, Ben Peters, Ricardo Rei, Patrick Fernandes, Sweta Agrawal, Pierre Colombo, José G.~C. de~Souza, and André F.~T. Martins. 2024.
\newblock \href {http://arxiv.org/abs/2402.17733} {Tower: An open multilingual large language model for translation-related tasks}.

\bibitem[{Carpuat et~al.(2024)Carpuat, Federico, Waibel, Niehues, St\"uker, Salesky, and Ojha}]{iwslt:2024}
Marine Carpuat, Marcello Federico, Alex Waibel, Jan Niehues, Sebastian St\"uker, Elizabeth Salesky, and Atul~Kr. Ojha. 2024.
\newblock Findings of the {IWSLT} 2024 {Evaluation} {Campaign}.
\newblock In \emph{Proceedings of the 21st {International} {Conference} on {Spoken} {Language} {Translation} ({IWSLT} 2024)}. Association for Computational Linguistics.

\bibitem[{Chen et~al.(2021)Chen, Chai, Wang, Du, Zhang, Weng, Su, Povey, Trmal, Zhang, Jin, Khudanpur, Watanabe, Zhao, Zou, Li, Yao, Wang, You, and Yan}]{gigaspeech}
Guoguo Chen, Shuzhou Chai, Guan-Bo Wang, Jiayu Du, Wei-Qiang Zhang, Chao Weng, Dan Su, Daniel Povey, Jan Trmal, Junbo Zhang, Mingjie Jin, Sanjeev Khudanpur, Shinji Watanabe, Shuaijiang Zhao, Wei Zou, Xiangang Li, Xuchen Yao, Yongqing Wang, Zhao You, and Zhiyong Yan. 2021.
\newblock \href {https://doi.org/10.21437/Interspeech.2021-1965} {{GigaSpeech: An Evolving, Multi-Domain ASR Corpus with 10,000 Hours of Transcribed Audio}}.
\newblock In \emph{Proc. Interspeech 2021}, pages 3670--3674.

\bibitem[{Chen et~al.(2022)Chen, Wang, Chen, Wu, Liu, Chen, Li, Kanda, Yoshioka, Xiao, Wu, Zhou, Ren, Qian, Qian, Wu, Zeng, Yu, and Wei}]{Chen_2022}
Sanyuan Chen, Chengyi Wang, Zhengyang Chen, Yu~Wu, Shujie Liu, Zhuo Chen, Jinyu Li, Naoyuki Kanda, Takuya Yoshioka, Xiong Xiao, Jian Wu, Long Zhou, Shuo Ren, Yanmin Qian, Yao Qian, Jian Wu, Michael Zeng, Xiangzhan Yu, and Furu Wei. 2022.
\newblock \href {https://doi.org/10.1109/jstsp.2022.3188113} {Wavlm: Large-scale self-supervised pre-training for full stack speech processing}.
\newblock \emph{IEEE Journal of Selected Topics in Signal Processing}, 16(6):1505–1518.

\bibitem[{Di~Gangi et~al.(2019)Di~Gangi, Cattoni, Bentivogli, Negri, and Turchi}]{di-gangi-etal-2019-must}
Mattia~A. Di~Gangi, Roldano Cattoni, Luisa Bentivogli, Matteo Negri, and Marco Turchi. 2019.
\newblock \href {https://doi.org/10.18653/v1/N19-1202} {{M}u{ST}-{C}: a {M}ultilingual {S}peech {T}ranslation {C}orpus}.
\newblock In \emph{Proceedings of the 2019 Conference of the North {A}merican Chapter of the Association for Computational Linguistics: Human Language Technologies, Volume 1 (Long and Short Papers)}, pages 2012--2017, Minneapolis, Minnesota. Association for Computational Linguistics.

\bibitem[{Huang et~al.(2023)Huang, Ye, Ko, Dong, Cheng, Wang, and Li}]{huang2023speech}
Zhichao Huang, Rong Ye, Tom Ko, Qianqian Dong, Shanbo Cheng, Mingxuan Wang, and Hang Li. 2023.
\newblock \href {http://arxiv.org/abs/2312.13585} {Speech translation with large language models: An industrial practice}.

\bibitem[{{Kahn} et~al.(2020){Kahn}, {Rivière}, {Zheng}, {Kharitonov}, {Xu}, {Mazaré}, {Karadayi}, {Liptchinsky}, {Collobert}, {Fuegen}, {Likhomanenko}, {Synnaeve}, {Joulin}, {Mohamed}, and {Dupoux}}]{librilight}
J.~{Kahn}, M.~{Rivière}, W.~{Zheng}, E.~{Kharitonov}, Q.~{Xu}, P.~E. {Mazaré}, J.~{Karadayi}, V.~{Liptchinsky}, R.~{Collobert}, C.~{Fuegen}, T.~{Likhomanenko}, G.~{Synnaeve}, A.~{Joulin}, A.~{Mohamed}, and E.~{Dupoux}. 2020.
\newblock Libri-light: A benchmark for asr with limited or no supervision.
\newblock In \emph{ICASSP 2020 - 2020 IEEE International Conference on Acoustics, Speech and Signal Processing (ICASSP)}, pages 7669--7673.
\newblock \url{https://github.com/facebookresearch/libri-light}.

\bibitem[{Ma et~al.(2019)Ma, Huang, Xiong, Zheng, Liu, Zheng, Zhang, He, Liu, Li, Wu, and Wang}]{ma-etal-2019-stacl}
Mingbo Ma, Liang Huang, Hao Xiong, Renjie Zheng, Kaibo Liu, Baigong Zheng, Chuanqiang Zhang, Zhongjun He, Hairong Liu, Xing Li, Hua Wu, and Haifeng Wang. 2019.
\newblock \href {https://doi.org/10.18653/v1/P19-1289} {{STACL}: Simultaneous translation with implicit anticipation and controllable latency using prefix-to-prefix framework}.
\newblock In \emph{Proceedings of the 57th Annual Meeting of the Association for Computational Linguistics}, pages 3025--3036, Florence, Italy. Association for Computational Linguistics.

\bibitem[{Ma et~al.(2020{\natexlab{a}})Ma, Dousti, Wang, Gu, and Pino}]{ma-etal-2020-simuleval}
Xutai Ma, Mohammad~Javad Dousti, Changhan Wang, Jiatao Gu, and Juan Pino. 2020{\natexlab{a}}.
\newblock \href {https://doi.org/10.18653/v1/2020.emnlp-demos.19} {{SIMULEVAL}: An evaluation toolkit for simultaneous translation}.
\newblock In \emph{Proceedings of the 2020 Conference on Empirical Methods in Natural Language Processing: System Demonstrations}, pages 144--150, Online. Association for Computational Linguistics.

\bibitem[{Ma et~al.(2020{\natexlab{b}})Ma, Dousti, Wang, Gu, and Pino}]{simuleval2020}
Xutai Ma, Mohammad~Javad Dousti, Changhan Wang, Jiatao Gu, and Juan Pino. 2020{\natexlab{b}}.
\newblock Simuleval: An evaluation toolkit for simultaneous translation.
\newblock In \emph{Proceedings of the EMNLP}.

\bibitem[{Nystrom et~al.(2015)Nystrom, Levine, Roskies, and Scott}]{nystrom2015bridges}
Nicholas~A Nystrom, Michael~J Levine, Ralph~Z Roskies, and J~Ray Scott. 2015.
\newblock Bridges: a uniquely flexible hpc resource for new communities and data analytics.
\newblock In \emph{Proceedings of the 2015 XSEDE Conference: Scientific Advancements Enabled by Enhanced Cyberinfrastructure}, pages 1--8.

\bibitem[{Ouyang et~al.(2022)Ouyang, Ye, and Li}]{ouyang-etal-2022-impact}
Siqi Ouyang, Rong Ye, and Lei Li. 2022.
\newblock \href {https://doi.org/10.18653/v1/2022.iwslt-1.9} {On the impact of noises in crowd-sourced data for speech translation}.
\newblock In \emph{Proceedings of the 19th International Conference on Spoken Language Translation (IWSLT 2022)}, pages 92--97, Dublin, Ireland (in-person and online). Association for Computational Linguistics.

\bibitem[{Papi et~al.(2022)Papi, Gaido, Negri, and Turchi}]{papi2022over}
Sara Papi, Marco Gaido, Matteo Negri, and Marco Turchi. 2022.
\newblock Over-generation cannot be rewarded: Length-adaptive average lagging for simultaneous speech translation.
\newblock In \emph{Proceedings of the Third Workshop on Automatic Simultaneous Translation}, pages 12--17.

\bibitem[{Papi et~al.(2023)Papi, Turchi, and Negri}]{Papi_2023}
Sara Papi, Marco Turchi, and Matteo Negri. 2023.
\newblock \href {https://doi.org/10.21437/interspeech.2023-170} {Alignatt: Using attention-based audio-translation alignments as a guide for simultaneous speech translation}.
\newblock In \emph{INTERSPEECH 2023}, interspeech\_2023. ISCA.

\bibitem[{Post(2018)}]{post-2018-call}
Matt Post. 2018.
\newblock \href {https://doi.org/10.18653/v1/W18-6319} {A call for clarity in reporting {BLEU} scores}.
\newblock In \emph{Proceedings of the Third Conference on Machine Translation: Research Papers}, pages 186--191, Brussels, Belgium. Association for Computational Linguistics.

\bibitem[{Touvron et~al.(2023)Touvron, Martin, Stone, Albert, Almahairi, Babaei, Bashlykov, Batra, Bhargava, Bhosale, Bikel, Blecher, Ferrer, Chen, Cucurull, Esiobu, Fernandes, Fu, Fu, Fuller, Gao, Goswami, Goyal, Hartshorn, Hosseini, Hou, Inan, Kardas, Kerkez, Khabsa, Kloumann, Korenev, Koura, Lachaux, Lavril, Lee, Liskovich, Lu, Mao, Martinet, Mihaylov, Mishra, Molybog, Nie, Poulton, Reizenstein, Rungta, Saladi, Schelten, Silva, Smith, Subramanian, Tan, Tang, Taylor, Williams, Kuan, Xu, Yan, Zarov, Zhang, Fan, Kambadur, Narang, Rodriguez, Stojnic, Edunov, and Scialom}]{touvron2023llama}
Hugo Touvron, Louis Martin, Kevin Stone, Peter Albert, Amjad Almahairi, Yasmine Babaei, Nikolay Bashlykov, Soumya Batra, Prajjwal Bhargava, Shruti Bhosale, Dan Bikel, Lukas Blecher, Cristian~Canton Ferrer, Moya Chen, Guillem Cucurull, David Esiobu, Jude Fernandes, Jeremy Fu, Wenyin Fu, Brian Fuller, Cynthia Gao, Vedanuj Goswami, Naman Goyal, Anthony Hartshorn, Saghar Hosseini, Rui Hou, Hakan Inan, Marcin Kardas, Viktor Kerkez, Madian Khabsa, Isabel Kloumann, Artem Korenev, Punit~Singh Koura, Marie-Anne Lachaux, Thibaut Lavril, Jenya Lee, Diana Liskovich, Yinghai Lu, Yuning Mao, Xavier Martinet, Todor Mihaylov, Pushkar Mishra, Igor Molybog, Yixin Nie, Andrew Poulton, Jeremy Reizenstein, Rashi Rungta, Kalyan Saladi, Alan Schelten, Ruan Silva, Eric~Michael Smith, Ranjan Subramanian, Xiaoqing~Ellen Tan, Binh Tang, Ross Taylor, Adina Williams, Jian~Xiang Kuan, Puxin Xu, Zheng Yan, Iliyan Zarov, Yuchen Zhang, Angela Fan, Melanie Kambadur, Sharan Narang, Aurelien Rodriguez, Robert Stojnic, Sergey Edunov, and Thomas
  Scialom. 2023.
\newblock \href {http://arxiv.org/abs/2307.09288} {Llama 2: Open foundation and fine-tuned chat models}.

\bibitem[{Towns et~al.(2014)Towns, Cockerill, Dahan, Foster, Gaither, Grimshaw, Hazlewood, Lathrop, Lifka, Peterson, Roskies, Scott, and Wilkins-Diehr}]{xsede}
J.~Towns, T.~Cockerill, M.~Dahan, I.~Foster, K.~Gaither, A.~Grimshaw, V.~Hazlewood, S.~Lathrop, D.~Lifka, G.~D. Peterson, R.~Roskies, J.~R. Scott, and N.~Wilkins-Diehr. 2014.
\newblock \href {https://doi.org/10.1109/MCSE.2014.80} {Xsede: Accelerating scientific discovery}.
\newblock \emph{Computing in Science \& Engineering}, 16(5):62--74.

\bibitem[{Wang et~al.(2021)Wang, Riviere, Lee, Wu, Talnikar, Haziza, Williamson, Pino, and Dupoux}]{wang-etal-2021-voxpopuli}
Changhan Wang, Morgane Riviere, Ann Lee, Anne Wu, Chaitanya Talnikar, Daniel Haziza, Mary Williamson, Juan Pino, and Emmanuel Dupoux. 2021.
\newblock \href {https://doi.org/10.18653/v1/2021.acl-long.80} {{V}ox{P}opuli: A large-scale multilingual speech corpus for representation learning, semi-supervised learning and interpretation}.
\newblock In \emph{Proceedings of the 59th Annual Meeting of the Association for Computational Linguistics and the 11th International Joint Conference on Natural Language Processing (Volume 1: Long Papers)}, pages 993--1003, Online. Association for Computational Linguistics.

\bibitem[{Zhang et~al.(2023)Zhang, Si, Chen, Zhang, Yang, Qu, and Jiao}]{zhang2023tuning}
Hao Zhang, Nianwen Si, Yaqi Chen, Wenlin Zhang, Xukui Yang, Dan Qu, and Xiaolin Jiao. 2023.
\newblock \href {http://arxiv.org/abs/2310.02050} {Tuning large language model for end-to-end speech translation}.

\end{thebibliography}

\end{document}